\documentclass{article}

\PassOptionsToPackage{numbers, compress}{natbib}

\usepackage[final]{tgl23}

\usepackage{booktabs}            % professional-quality tables
\usepackage{multirow}            % tabular cells spanning multiple rows
\usepackage{amsfonts}            % blackboard math symbols
\usepackage{graphicx}            % figures
\usepackage{duckuments}          % sample images
\usepackage{subcaption}
\usepackage{makecell}

\usepackage{amsmath} % for \boldsymbol macro
\usepackage{bm}      % for \bm macro
\usepackage{dsfont}
\usepackage{amsfonts,amssymb,stmaryrd}
\usepackage{ulem}
\useunder{\uline}{\ul}{} % shortcut to underline in table 3

\usepackage[utf8]{inputenc} % allow utf-8 input
\usepackage[T1]{fontenc}    % use 8-bit T1 fonts
\usepackage{hyperref}       % hyperlinks
\usepackage{url}            % simple URL typesetting
\usepackage{booktabs}       % professional-quality tables
\usepackage{amsfonts}       % blackboard math symbols
\usepackage{nicefrac}       % compact symbols for 1/2, etc.
\usepackage{microtype}      % microtypography
\usepackage{xcolor}         % colors
\usepackage{amssymb}
\usepackage{amsmath}
\usepackage{caption}
\usepackage{subcaption}
\usepackage{graphicx}
\usepackage{multirow}
\usepackage{verbatim}
\usepackage{multirow}            % tabular cells spanning multiple rows
\usepackage{makecell}
\usepackage{bm}      % for \bm macro
\usepackage{dsfont}
\usepackage{amsfonts,amssymb,stmaryrd}
\usepackage{ulem}
\usepackage{cleveref}

\newcommand{\opnorm}[1]{{\left\vert\kern-0.25ex\left\vert\kern-0.25ex\left\vert #1 
    \right\vert\kern-0.25ex\right\vert\kern-0.25ex\right\vert}}

\newcommand\restr[2]{{% we make the whole thing an ordinary symbol
  \left.\kern-\nulldelimiterspace % automatically resize the bar with \right
  #1 % the function
  \vphantom{\big|} % pretend it's a little taller at normal size
  \right|_{#2} % this is the delimiter
  }}

\usepackage{xspace}

\newcommand*{\ie}{i.e.\@\xspace}

\newcommand{\norm}[1]{\left\lVert#1\right\rVert}

\author{%
  Manuel Dileo$^1$, Pasquale Minervini$^2$, Matteo Zignani$^1$, Sabrina Gaito$^1$\\
  $^1$ Department of Computer Science, University of Milan, Milan, Italy\\
  $^2$ School of Informatics, University of Edinburgh, Edinburgh, United Kingdom\\
  \texttt{manuel.dileo@unimi.it}\\
}

\title{Temporal Smoothness Regularisers for Neural Link Predictors}

\begin{document}

\maketitle

\begin{abstract}
Most algorithms for representation learning and link prediction on relational data are designed for static data.
However, the data to which they are applied typically evolves over time, including online social networks or interactions between users and items in recommender systems.
This is also the case for graph-structured knowledge bases -- knowledge graphs -- which contain facts that are valid only for specific points in time. In such contexts, it becomes crucial to correctly identify missing links at a precise time point, i.e. the temporal prediction link task.
Recently, \citet{DBLP:conf/iclr/LacroixOU20} and \citet{DBLP:conf/aaai/SadeghianACW21} proposed a solution to the problem of link prediction for knowledge graphs under temporal constraints inspired by the canonical decomposition of $4$-order tensors, where they regularise the representations of time steps by enforcing temporal smoothing, i.e. by learning similar transformation for adjacent timestamps. However, the impact of the choice of temporal regularisation terms is still poorly understood.
In this work, we systematically analyse several choices of temporal smoothing regularisers using linear functions and recurrent architectures.
In our experiments, we show that by carefully selecting the temporal smoothing regulariser and regularisation weight, a simple method like TNTComplEx~\cite{DBLP:conf/iclr/LacroixOU20} can produce significantly more accurate results than state-of-the-art methods on three widely used temporal link prediction datasets.
Furthermore, we evaluate the impact of a wide range of temporal smoothing regularisers on two state-of-the-art temporal link prediction models.
We observe that linear regularisers for temporal smoothing based on specific nuclear norms can significantly improve the predictive accuracy of the base temporal link prediction methods. 
Our work shows that simple tensor factorisation models can produce new state-of-the-art results using newly proposed temporal regularisers, highlighting a promising avenue for future research.
%with nuclear N5 norm achieve the best performances.
%
\end{abstract}

\section{Introduction}

Knowledge Graphs (KGs)~\citep{DBLP:journals/csur/HoganBCdMGKGNNN21} are graph-structured knowledge bases where knowledge about the world is encoded in the form of relationships of various kinds between entities.
KGs are an extremely flexible and versatile knowledge representation formalism, and are currently being used in a variety of domains, including bioinformatics~\citep{DBLP:journals/bioinformatics/ZitnikAL18,DBLP:journals/corr/abs-2305-19979,DBLP:journals/bib/MohamedNN21,DBLP:conf/cikm/WalshMN20,Himmelstein087619}, recommendation~\citep{DBLP:journals/tkde/GuoZQZXXH22}, linguistics~\citep{DBLP:conf/naacl/Miller92}, and industry applications~\citep{DBLP:journals/cacm/NoyGJNPT19}.
Moreover, relational data from such domains is often temporal; for example, the action of buying an item or watching a movie is associated with a timestamp, and some medicines might interact differently depending on when they are administered to a patient. We refer to KGs augmented with temporal information as \emph{Temporal Knowledge Graphs} (TKGs). 

When reasoning with TKGs, one of the most crucial tasks is finding or completing missing links in the temporal knowledge graph at a precise time point, often referred to as the \emph{temporal link prediction} task.
%
% The task of \emph{temporal link prediction} consists in \emph{finding missing links in graphs at precise points in time}.
%
%\emph{Neural link predictors}~\citep{DBLP:journals/pieee/Nickel0TG16} tackle the problem of identifying missing edges in large static or temporal KGs.
One class of models for tackling the problem of identifying missing links in large static or temporal KGs is \emph{neural link predictors}~\citep{DBLP:journals/pieee/Nickel0TG16}, \ie differentiable models which map entities and relationships into a $d$-dimensional embedding space, and use entity and relation embeddings for scoring missing links in the graph.

%
%However, relational data is often temporal; for example, the action of buying an item or watching a movie is associated with a timestamp, and some medicines might interact differently with each other depending on when in time they are administered to a patient.
%
%The task of \emph{temporal link prediction} consists in \emph{finding missing links in graphs at precise points in time}.
%
Recently, \citet{DBLP:conf/iclr/LacroixOU20} and \citet{DBLP:conf/aaai/SadeghianACW21} proposed two state-of-the-art approaches to temporal link prediction for temporal knowledge graphs. The key idea is to extend factorisation-based neural link prediction models with dense representations for each timestamp and to regularise such representations by enforcing them to change slowly over time.
However, the impact of the choice of temporal regularisation terms is still not well understood.

Hence, in this work, we systematically analyse a comprehensive array of temporal regularisers for neural link predictors based on tensor factorisation. Starting from the temporal smoothing regulariser proposed in \cite{DBLP:conf/iclr/LacroixOU20}, we also consider a wide class of norms in the L$_{p}$ and N$_{p}$ families which allow to control the strength of the smoothing. We further extend the set of temporal regulariser by taking into account the regulariser proposed in \cite{DBLP:conf/naacl/XuCNL21}, defining it as an explicit modelling of the temporal dynamic between adjacent timestamps. Lastly, we propose to adopt a recurrent architecture as an implicit temporal regulariser that can generate timestamp embeddings sequentially and learn the temporal dynamic by updating its parameters during the training phase.

We conducted the experimental evaluation over three well-known benchmark datasets: ICEWS14, ICEWS05-15, and YAGO15K.
ICEWS14 and ICEWS05-15 are subsets of the widely used Integrated Crisis Early Warning System knowledge graph \cite[ICEWS,][]{icewsdataset}, while YAGO15K~\cite{DBLP:conf/emnlp/Garcia-DuranDN18} is a dataset derived from YAGO~\cite{DBLP:conf/semweb/RebeleSHBKW16} covering the entities appearing in FB15k \cite{DBLP:conf/nips/BordesUGWY13}, which adds \textit{occursSince} and \textit{occursUntil} timestamps to each triple.

Our results show that using our proposed temporal regularisers, neural link predictors based on tensor factorisation models can significantly improve their predictive accuracy in temporal link prediction tasks.
By carefully selecting a temporal regulariser and regularisation weight, our version of TNTComplEx \cite{DBLP:conf/iclr/LacroixOU20} produces more accurate results than all of the baselines on ICEWS14, ICEW05-15, and YAGO15K.
Overall, linear regularizers for temporal smoothing that introduce smaller loss penalties for closer timestamp representations achieve the best performance.
In contrast, recurrent architecture struggles to generate a long sequence of timestamps.

%The paper is organized as follows. 
%Section \ref{sec:background} reviews several works related to knowledge graph completion for both static and temporal data. In Section \ref{sec:method} we present a framework for temporal knowledge graph representation learning, and we describe the models and temporal regularisers considered for our analysis. In Section \ref{sec:experiments}, we provide a description of the dataset, evaluation setup and a discussion on the results, while Section \ref{sec:conclusion} resumes the main findings and suggests a potential direction for future research.

% Also talk about adaptive regulatisers and RNN-based architecture (should we talk about it tho?)
%
% Shall we invite Yihong to this overleaf tho? She's great!
%

%
\section{Background}
\label{sec:background}
A Knowledge Graph $\mathcal{G} \subseteq \mathcal{E} \times \mathcal{R} \times \mathcal{E}$ contains a set of subject-predicate-object $\langle s, p, o \rangle$ triples, where each triple represents a relationship of type $p \in \mathcal{R}$ between the subject $s \in \mathcal{E}$ and the object $o \in \mathcal{E}$ of the triple.
Here, $\mathcal{E}$ and $\mathcal{R}$ denote the set of all entities and relation types, respectively.
However, many real-world knowledge graphs are largely incomplete~\citep{DBLP:conf/aaai/DettmersMS018,DBLP:journals/semweb/FarberBMR18,DBLP:journals/pieee/Nickel0TG16,DBLP:journals/ivs/DestandauF21} -- link prediction focuses on the problem of identifying missing links in (possibly very large) knowledge graphs.
\subsection{Static Neural Link Predictors}
\emph{Neural link predictors}, also referred to as KG embedding models, are neural models that yield state-of-the-art accuracy on a wide array of link prediction benchmarks~\citep{DBLP:conf/iclr/RuffinelliBG20,DBLP:conf/rep4nlp/KadlecBK17,DBLP:journals/corr/abs-2005-00804}.
A neural link predictor is a neural model where entities in $\mathcal{E}$ and relation types in $\mathcal{R}$ are represented in a continuous embedding space, and the likelihood of a link between two entities is a function of their representations.
More formally, neural link predictors are defined by a parametric \emph{scoring function} $\phi_{\theta}: \mathcal{E} \times \mathcal{R} \times \mathcal{E} \mapsto \mathbb{R}$, with parameters $\theta$ that, given a triple $\langle s, p, o \rangle$, produces the likelihood that entities $s$ and $o$ are related by the relationship $p$.
See Appendix for a detailed description of static neural link predictors.
\subsection{Temporal Knowledge Graph Completion}

A \emph{Temporal Knowledge Graph} (TKG) is referred to a set of quadruples $\mathcal{K} \subseteq \{(s, p, o, \tau) \mid s,o \in \mathcal{E}, p \in \mathcal{R}, \tau \in \mathcal{T} \}$. Each quadruple represents a temporal fact that is true in a world. $\mathcal{E}$ is the set of all entities and $\mathcal{R}$ is the set of all relations in the ontology. The fourth element in each quadruple represents time, which is often discretised. $\mathcal{T}$ represents the set of all possible timestamps. Temporal Knowledge Graph Completion, also known as temporal link prediction, refers to the problem of completing a TKG by inferring facts from a given subset of its facts.

The subject of temporal link prediction has been studied in a wide range of approaches~\cite{DBLP:journals/jmlr/KazemiGJKSFP20}. For a general overview of temporal link prediction models see the Appendix. In this work, we focus on methods that learn the temporal behaviour by using a representation for time. For instance, \citet{DBLP:conf/iclr/LacroixOU20} perform tensor decomposition based on the time representation, while \citet{DBLP:conf/aaai/SadeghianACW21} learns a $d$-dimensional rotation transformation parametrized by relation and time, such that after each fact's head entity representation is transformed using the rotation, it falls near its corresponding tail entity. These methods achieve state-of-the-art results on most of the benchmark datasets~\cite{DBLP:conf/aaai/SadeghianACW21, DBLP:conf/naacl/XuCNL21}. Hence, we systematically analyse a wide array of temporal regularisers to understand their impact on both performance and learning temporal behaviour.S

\section{Temporal KG Representation Learning}
\label{sec:method}
This section presents a framework for temporal knowledge graph representation learning \cite{DBLP:conf/iclr/LacroixOU20}.
Given a TKG, we want to learn representations for entities, relations, and timestamps (e.g., $\bf{s}, \bf{p}, \bf{o}, \bf{t_\tau} \in \mathbb{R}^{d}$) and a scoring function $\phi_{\theta}(s,p,o,\tau) \in \mathbb{R}$, such that true quadruples receive high scores. Thus, given $\phi_{\theta}$, the embeddings can be learned by optimising an appropriate cost function.
Following \citet{DBLP:conf/icml/LacroixUO18}, we minimise, for each of the train tuples $(s,p,o,\tau)$, the instantaneous multi-class loss:
\begin{equation}
    \ell(\phi_{\theta};(s,p,o,\tau)) = - \phi_{\theta}(s,p,o,\tau) + \log\left(\sum_{o^{\prime} \in \mathcal{E}}\exp\left(\phi_{\theta}(s,p,o^{\prime},\tau)\right)\right),
    \label{eq:loss}
\end{equation}
%
%\noindent where $k^{\prime} \in \mathcal{E}$ are all the entities not involved in positive triples with $i$.
\noindent Note that this loss is only suited to queries of the type (subject, predicate, ?, time), which are the queries that were considered in related work.
For a training set $S$ (augmented with reciprocal relations \citep{DBLP:conf/icml/LacroixUO18, DBLP:conf/nips/Kazemi018}), and parametric tensor estimate $\phi_{\theta}$, we minimize the following objective, with a weighted embedding regulariser $\Omega$:
\begin{equation}
    \mathcal{L}(\phi_{\theta}) = \frac{1}{|S|}\sum_{(s,p,o,\tau)\in S}\left[\ell(\phi_{\theta};(s,p,o,\tau)) + \lambda\Omega(\theta; (s,p,o,\tau))\right].
    \label{loss-with-reg}
\end{equation}
In our experiments, embedding regularisation is performed using a nuclear tensor 3-norm \cite{DBLP:conf/icml/LacroixUO18}.
In the following subsections, we introduce the models, scoring functions, and temporal regularisers considered for our analysis.

\subsection{TNTComplEx}
TNTComplEx \cite{DBLP:conf/iclr/LacroixOU20} extends the ComplEx \cite{DBLP:conf/icml/LacroixUO18} decomposition to the TKG completion setting by adding a new factor $T$ resulting in the following scoring function $\phi_{\theta}$:
% \begin{equation}
%     \empx(U, V, T) = \operatorname{Re}\left(\brck{U,V,\overline{U}, T} \right)\iff \empx(U,V,T)_{i,j,k,l} = \operatorname{Re}\left(\langle \bf{i}, \bf{j}, \overline{\bf{k}}, \bf{t_l}\rangle\right)
%     \label{eq:tntcomplex1}
% \end{equation}
\begin{equation} \label{eq:tntcomplex1}
\phi_{\theta}^{\text{TComplEx}}(s,p,o,\tau) = \operatorname{Re}\left(\langle \bf{s}, \bf{p}, \overline{\bf{o}}, \bf{t_\tau}\rangle\right)
\end{equation}
where $\bf{s}$ and $\bf{o}$ are the embeddings of entities $s,o \in \mathcal{E}$, $\bf{p}$ is the embedding for the relation $p \in \mathcal{R}$, and $\bf{t_\tau}$ is the embedding for the timestamp $\tau \in \mathcal{T}$.
Intuitively, they added timestamps embedding that modulate the multi-linear dot product.

In heterogeneous knowledge bases, where only part of the relation types are temporal, they introduce an embedding representation -- $\bf{p}^t$ -- whether the relation $p$ is temporal and an embedding $\bf{p}$ otherwise. Thus, the scoring function $\phi_{\theta}^{\text{TComplEx}}$ is extended into the TNTComplEx scoring function $\phi_{\theta}^{\text{TNTComplEx}}$:
%
% \begin{equation}
%     \empx = \operatorname{Re}\left(\brck{U,V^t,\overline{U}, T} + \brck{U,V,\overline{U}, \mathbf{1}}\right)\iff \empx_{i,j,k,l} = Re\left(\langle \bf{i}, \bf{j^t}\odot \bf{t_l} + \bf{j}, \overline{\bf{k}}\rangle\right)
%     \label{eq:tntcomplex2}
% \end{equation}
% where $V^t$ denotes the embedding matrix for temporal predicates and $V$ the embedding matrix for non-temporal predicates.
\begin{equation}\label{eq:tntcomplex2}
    \phi_{\theta}^{\text{TNTComplEx}}(s,p,o,\tau) = Re\left(\langle \bf{s}, \bf{p}^t\odot \bf{t_\tau} + \bf{p}, \overline{\bf{o}}\rangle\right)
\end{equation}

\subsection{ChronoR}
ChronoR~\cite{DBLP:conf/aaai/SadeghianACW21} is inspired by rotation-based models in static KG completion~\cite{DBLP:conf/iclr/SunDNT19}. For a training set $S$ containing train tuples $(s,p,o,\tau)$, with $|S| = n$, they expect that for true facts it holds:
\begin{equation}
    \operatorname{Q}_{\bf{p}, \bf{t_\tau}}\odot \;\bf{s} = \bf{o}
    \label{eq:rotate}
\end{equation}
\noindent where %, by abuse of notation, 
$\bf{s}, \bf{o} \in \mathbb{R}^{n \times d}$ denote the entity embeddings obtained by vertically concatenating the embedding of the entity $s$ or $o$ in each training tuple, and $\operatorname{Q}_{\bf{p}, \bf{t_\tau}}$ represents the (row-wise) linear operator parameterised by the relation and timestamp embeddings $\bf{p}$ and $\bf{t_\tau}$. Specifically, $\operatorname{Q}$ is parameterized by $\bf{p}, \bf{t_\tau}$ by concatenating the embeddings to get $\operatorname{Q}_{\bf{p},\bf{t_\tau}} = [\;\bf{p}\;|\; \bf{t_\tau}\;]$, where $\bf{p} \in \mathbb{R}^{n_p \times d}$ and $\bf{t_\tau} \in \mathbb{R}^{n_\tau \times d}$ are the representations of the relation type and time elements, and $n_p + n_\tau = n$. Thus, the scoring function of ChronoR is defined as:
\begin{equation}
\label{eq:ChronoR1}
    \phi_{\theta}^{\text{ChronoR}}(s, p, o, \tau) = \left\langle \bf{s} \, \odot [\;\bf{p}\;|\;\bf{t_\tau}\;]_{r,:}, \bf{o} \right\rangle,
\end{equation}
\noindent where $r$ indicates the row in $\operatorname{Q}_{\bf{p}, \bf{t_\tau}}$ corresponding to the train tuple $(s, p, o, \tau)$. It is worth noting that, unlike RotatE \cite{DBLP:conf/iclr/SunDNT19}, which uses a scoring function based on the Euclidean distance of $\operatorname{Q}_{\bf{p}}\odot\;\bf{s}$ and $\bf{o}$, they propose to use the angle between the two vectors, or the Frobenius inner product for the matrix formulation.

As well as for TNTComplEx, to treat TKGs that store a combination of static and dynamic facts, they allow an extra rotation operator $\bf{p}'\in \mathbb{R}^{n \times d}$, leading to the following scoring function:
\begin{equation}
\label{eq:ChronoR2}
    \phi_{\theta}^{\text{ChronoR}}(s, p, o, \tau) = \left\langle \bf{s} \, \odot [\;\bf{p}\;|\;\bf{t_\tau}\;]_{r,:} \odot \bf{p'}_{r,:}, \bf{o} \right\rangle.
\end{equation}
% where $\bf{j'} \in \mathbb{R}^{n \times m}$.

\subsection{Temporal regularisers}
Temporal regularisers encourage tensor factorisation models to learn transformation for timestamp embeddings that capture specific temporal properties of real datasets. For instance, one would like the model to take advantage of the fact that most entities behave smoothly over time, \ie learn similar transformations for closer timestamps. Alternatively, one would like to push away the representation of distant timestamps or allow the timestamp embeddings to be generated sequentially. 

Formally, a temporal regulariser is a weighted penalty term $\Lambda(T)$ in the loss function, leading to the minimisation of the following objective:
\begin{equation}
    \mathcal{L}(\phi_{\theta}) = \frac{1}{|S|}\sum_{(s,p,o,\tau)\in S}\left[\ell(\phi_{\theta};(s,p,o,\tau)) + \lambda_1\Omega(\theta; (s,p,o,\tau))\right] + \lambda_2 \Lambda(T) .
    \label{eq:loss-with-tempreg}
\end{equation}
Below we define the temporal regularisers considered in our analysis.

\paragraph{Temporal smoothing} 

\begin{figure}
    \centering
    \includegraphics[scale=0.25]{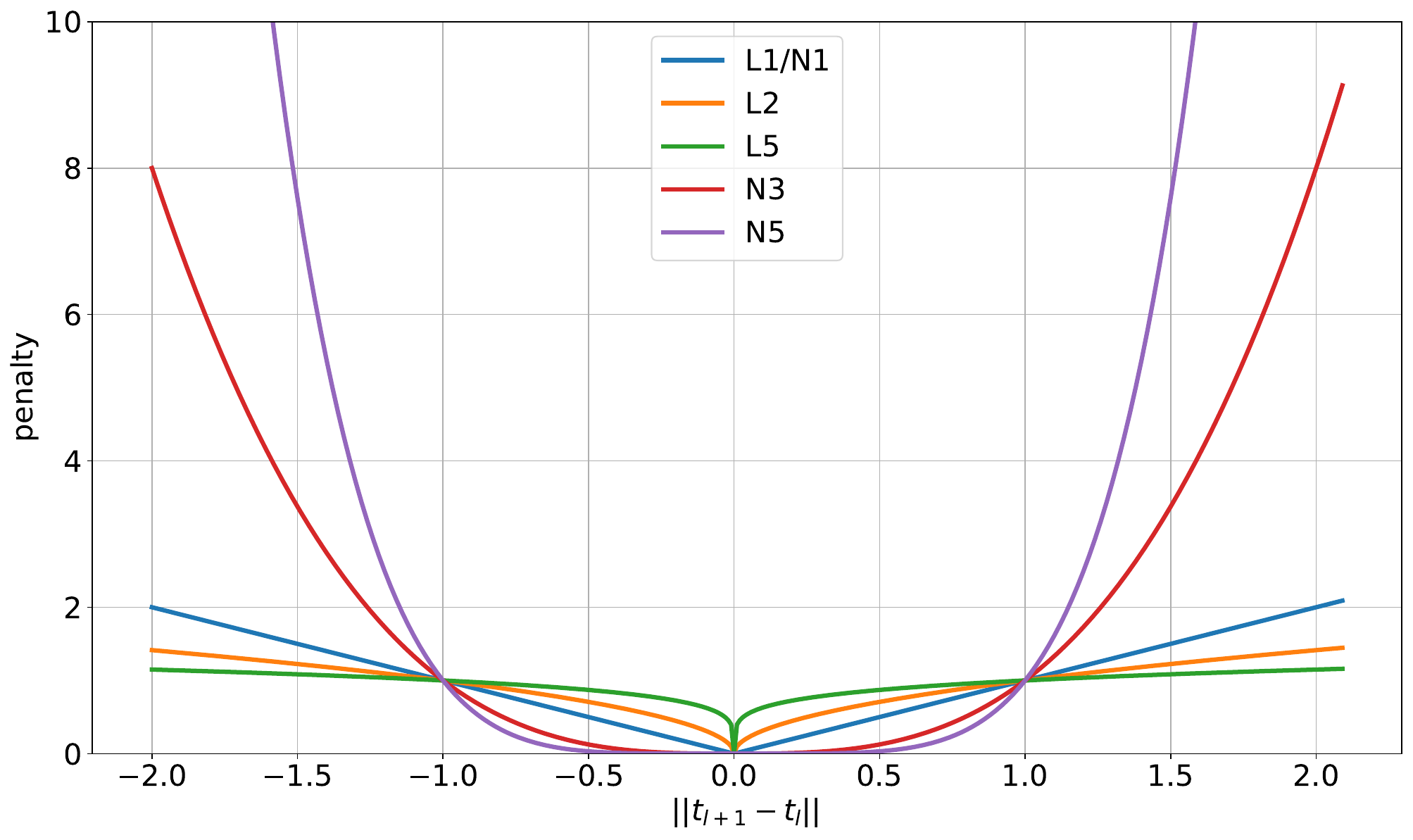}
    \caption{Plot of norm functions on the interval [-2; 2].}
    \label{fig:norms}
\end{figure}

Most of the work in the literature \cite{DBLP:conf/iclr/LacroixOU20,DBLP:conf/aaai/SadeghianACW21,DBLP:conf/aaai/MessnerAC22} adds a temporal smoothness objective to the loss function to encourage neighbouring timestamps to have close representations. The temporal smoothing regulariser can be defined as:
\begin{equation}
    \Lambda_p(T) = \frac{1}{|T| - 1}\sum_{\tau=1}^{|T|-1}\|\bf{t_{\tau+1}} - \bf{t_\tau}\|_p^p.
    \label{eq:temporal-smoothing}
\end{equation}
Despite being a very simple temporal regulariser, smoothing leads to an increase in performance on several benchmark datasets \cite{DBLP:conf/iclr/LacroixOU20,DBLP:conf/aaai/SadeghianACW21}.
However, it has been tested using only the N$_{3}$ norm. Hence, we decide to further investigate its impact by considering a wide class of norms in the L$_{p}$ and N$_{p}$ families, leading to the following temporal regularisation terms:
\begin{align}
%\begin{aligned}
    \Lambda_{L_p}(T) & = \frac{1}{|T| - 1}(\sum_{\tau=1}^{|T|-1}|\bf{t_{\tau+1}} - \bf{t_\tau}|^p)^{\frac{1}{p}},
    \label{eq:temporal-smoothing-Lp} \\
%\end{equation}
%\begin{equation}
    \Lambda_{N_p}(T) & = \frac{1}{|T| - 1}\sum_{\tau=1}^{|T|-1}|\bf{t_{\tau+1}} - \bf{t_\tau}|^p.
    \label{eq:temporal-smoothing-Np}
%\end{aligned}
\end{align}
In our intuition, the choice of the norm function controls the strength of the smoothing.
Thus, it provides several ways to penalize a change in the behaviour of the entities on neighbouring timestamps.
We plot some of the considered norms in \cref{fig:norms}.
Using an L$_{p}$ norm always penalises no equal timestamp representations, and the value of $p$ determines the magnitude of the penalty. Using an N$_{p}$ norm allows not penalising close representation for neighbouring timestamps (in the figure, N$_{5}$ starts to grow only for $|\bf{t_{\tau+1}}-\bf{t_{\tau}}| \geq 0.4$). In this case, the value of $p$ controls the magnitude of the penalty and the range of distances to not penalise.

\paragraph{Linear3 Regulariser}

In \cite{DBLP:conf/naacl/XuCNL21}, they propose a new temporal regulariser, namely \emph{Linear3}, that can be defined as follows:
\begin{equation}
    \Lambda_p(T) = \frac{1}{|T| - 1}\sum_{\tau=1}^{|T|-1}\|\bf{t_{\tau+1}} - \bf{t_\tau} - \bf{W_b}\|_p^p,
    \label{eq:linear3}
\end{equation}
\noindent where $\bf{W_b} \in \mathbb{R}^d$ denotes the embedding of a bias component between the neighbouring temporal embeddings, $d$ the embedding size. The bias embedding is randomly initialised and then learned from the training process. 

This linear regulariser promotes that the difference between embeddings of two adjacent timestamps is smaller than the difference between embeddings of two distant timestamps. In the context of this study, it is interesting to notice that Linear3 can be interpreted as the average score of triples $(\tau+1, \; \mathrm{\textit{follows}},\; \tau)$, where the embedding of the relation \textit{follows} is given by the bias component. From this point of view, Linear3 encourages similar embeddings for neighbouring timestamps by \emph{explicitly} modelling their temporal dynamic through the predicate \textit{follows}.

%\paragraph{RNN Regulariser}
\paragraph{Modelling Temporal Dynamics via Recurrent Architectures}
Timestamp embeddings can be generated sequentially using a recurrent neural architecture that, starting from a random initialised hidden state, can learn \emph{implicitly} the temporal dynamic through the learning process.
Given a specific Recurrent Neural Network (RNN) architecture, the equation that describes its forward procedure acts as a temporal regulariser that maps the embedding of timestamp $\tau$ as:
\begin{equation}
    \bf{t_\tau} = \mathrm{\textit{MLP}}(\mathrm{\textit{RNN}}(h_{\tau-1}, \emph{\bm{0}})); \tau\in \{1,...,|\mathcal{T}|\}
    \label{eq:rnn-reg}
\end{equation}
\noindent where $h_0 \in \mathbb{R}^m$ is the learnable initial hidden state, $\emph{\bm{0}}$ is the zero vector, \textit{RNN} is the function that describes the recurrent architecture, \textit{MLP} is a function that describes one multi-layer perceptron layer that has $(m,d)$ channels, $d$ is the embedding dimension and $m < d$.

In our experiments, we consider RNN, LSTM, and GRU as recurrent architecture and their linear counterpart, \ie by %setting to the identity their 
removing non-linear operations from the model. %non-linearity functions.

\section{Experiments}
\label{sec:experiments}

We evaluate the impact of temporal regularisers on TNTComplEx and ChronoR models for temporal link prediction on temporal knowledge graphs. We tune all the hyper-parameters on the validation set provided with each dataset using a grid search. We tune $\lambda_1$ and $\lambda_2$ from $\{10^i| -4 \leq i \leq 1\}$, the $p$ value for N$_{p}$ and L$_{p}$ norms from $\{i | 1 \leq i \leq 5\}$, the embedding dimensions for tensors and recurrent architecture from $\{5, 25, 50, 100, 500, 2000\}$. 

The training was done using mini-batch stochastic gradient descent with Adam and a learning rate of 0.1 with a batch size of 1000 quadruples. We implemented all our models in PyTorch. The source code to reproduce the full experimental results is made public on GitHub\footnote{\url{https://anonymous.4open.science/r/tkbc-reg-B77C/}}.

\subsection{Datasets} \label{sec:Dataset}
We assess the performance of our model on three well-known benchmarks for Temporal Knowledge graph completion: ICEWS14, ICEWS05-15, and YAGO15K. All these datasets exclusively consist of positive triples. ICEWS14 and ICEWS05-15 are subsets of the widely used Integrated Crisis Early Warning System (ICEWS) knowledge graph. ICEWS14 covers the time span from 01/01/2014 to 12/31/2014, while ICEWS15-05 represents the subset between 01/01/2005 and 12/31/2015. Both datasets include timestamp information for each fact, with a temporal resolution of 24 hours. Importantly, these datasets are deliberately selected to encompass only the most frequently occurring entities in both the head and tail positions \cite{DBLP:conf/iclr/LacroixOU20, DBLP:conf/aaai/SadeghianACW21}.
To create YAGO15K, \citet{DBLP:conf/emnlp/Garcia-DuranDN18} aligned the entities in FB15K~\cite{DBLP:conf/nips/BordesUGWY13} with those from YAGO~\cite{DBLP:conf/semweb/RebeleSHBKW16}, which contains temporal information. The final dataset is the result of all facts with successful alignment. It is worth noting that since YAGO does not have temporal information for all facts, this dataset is also temporally incomplete and more challenging. To adapt YAGO15K to our models, following~\cite{DBLP:conf/iclr/LacroixOU20}, for each fact, we group the relations \textit{occursSince} and \textit{occursUntil} together, in turn doubling our relation size. Note that this does not affect the evaluation protocol. \cref{tab:stats}, summarises the statistics of used temporal KG benchmarks.

\begin{table}
\centering
\begin{tabular}{lccc}
\toprule
          & {\bf ICEWS14} & {\bf ICEWS05-15}  & {\bf YAGO15K}     \\ 
\midrule
\# Entities   & 7,128   & 10,488      & 15,403      \\
\# Relations  & 230     & 251         & 34          \\
\# Timestamps & 365     & 4,017       & 198         \\
\# Facts      & 90,730  & 479,329     & 138,056     \\
Time Spans  & 2014    & 2005 - 2015 & 1513 - 2017 \\ 
\bottomrule
\end{tabular}
\caption{Dataset statistics for ICEWS14, ICEWS05-15, and YAGO15K.}
\label{tab:stats}
\end{table}

\subsection{Evaluation setup}
We follow the experimental set-up described in~\cite{DBLP:conf/emnlp/Garcia-DuranDN18} and~\cite{DBLP:conf/aaai/GoelKBP20}. For each quadruple $(s, p, o, \tau)$ in the test set, we fill $(s,p,?,\tau)$ and $(?,p, o, \tau)$ by scoring and sorting all possible entities in $\mathcal{E}$. We report Hits@k for $k=1, 3, 10$ and filtered Mean Reciprocal Rank (MRR) for all datasets. We refer to \citet{DBLP:journals/pieee/Nickel0TG16} for more details about the evaluation metrics.

\begin{table*}[t]
\setlength{\tabcolsep}{4pt}
\resizebox{\textwidth}{!}{%
\begin{tabular}{lcccccccccccc}
\toprule
 &
  \multicolumn{4}{c}{{\bf ICEWS14}} &
  \multicolumn{4}{c}{{\bf ICEWS05-15}} &
  \multicolumn{4}{c}{{\bf YAGO15K}} \\

\cmidrule(lr){2-5}
\cmidrule(lr){6-9}
\cmidrule(lr){10-13}

\multicolumn{1}{l}{\bf Model} &
  MRR &
  Hit@1 &
  Hits@3 &
  \multicolumn{1}{c}{Hit@10} &
  MRR &
  Hit@1 &
  Hit@3 &
  \multicolumn{1}{c}{Hit@10} &
  MRR &
  Hit@1 &
  Hit@3 &
  Hit@10 \\
  \midrule
\multicolumn{1}{l}{TransE (2013)} &
  28.0 &
  9.4 &
  - &
  \multicolumn{1}{c}{63.70} &
  29.4 &
  8.4 &
  - &
  \multicolumn{1}{c}{66.30} &
  29.6 &
  22.8 &
  - &
  46.8 \\
\multicolumn{1}{l}{DistMult (2014)} &
  43.9 &
  32.3 &
  - &
  \multicolumn{1}{c}{67.2} &
  45.6 &
  33.7 &
  - &
  \multicolumn{1}{c}{69.1} &
  27.5 &
  21.5 &
  - &
  43.8 \\
\multicolumn{1}{l}{SimpIE (2018)} &
  45.8 &
  34.1 &
  51.6 &
  \multicolumn{1}{c}{68.7} &
  47.8 &
  35.9 &
  53.9 &
  \multicolumn{1}{c}{70.8} &
  - &
  - &
  - &
  - \\
\multicolumn{1}{l}{ComplEx (2016)} &
  47.0 &
  35.0 &
  54.0 &
  \multicolumn{1}{c}{71.0} &
  49.0 &
  37.0 &
  55.0 &
  \multicolumn{1}{c}{73.0} &
  36.0 &
  29.0 &
  36.0 &
  54.0 \\
  \midrule
\multicolumn{1}{l}{ConT (2018)} &
  18.5 &
  11.7 &
  20.5 &
  \multicolumn{1}{c}{31.50} &
  16.4 &
  10.5 &
  18.9 &
  \multicolumn{1}{c}{27.20} &
  - &
  - &
  - &
  - \\
\multicolumn{1}{l}{TTransE (2016)} &
  25.5 &
  7.4 &
  - &
  \multicolumn{1}{c}{60.1} &
  27.1 &
  8.4 &
  - &
  \multicolumn{1}{c}{61.6} &
  32.1 &
  23.0 &
  - &
  51.0 \\
\multicolumn{1}{l}{TA-TransE (2018)} &
  27.5 &
  9.5 &
  - &
  \multicolumn{1}{c}{62.5} &
  29.9 &
  9.6 &
  - &
  \multicolumn{1}{c}{66.8} &
  32.1 &
  23.1 &
  - &
  51.2 \\
\multicolumn{1}{l}{HyTE (2018)} &
  29.7 &
  10.8 &
  41.6 &
  \multicolumn{1}{c}{65.5} &
  31.6 &
  11.6 &
  44.5 &
  \multicolumn{1}{c}{68.1} &
  - &
  - &
  - &
  - \\
\multicolumn{1}{l}{TA-DistMult (2018)} &
  47.7 &
  - &
  36.3 &
  \multicolumn{1}{c}{68.6} &
  47.4 &
  34.6 &
  - &
  \multicolumn{1}{c}{72.8} &
  29.1 &
  21.6 &
  - &
  47.6 \\
\multicolumn{1}{l}{DE-SimpIE (2020)} &
  52.6 &
  41.8 &
  59.2 &
  \multicolumn{1}{c}{72.5} &
  51.3 &
  39.2 &
  57.8 &
  \multicolumn{1}{c}{74.8} &
  - &
  - &
  - &
  - \\
\multicolumn{1}{l}{TIMEPLEX (2020)} &
  60.40 &
  51.50 &
  - &
  \multicolumn{1}{c}{77.11} &
  63.99 &
  54.51 &
  - &
  \multicolumn{1}{c}{81.81} &
  - &
  - &
  - &
  - \\
\multicolumn{1}{l}{TeRo (2020)} &
  56.2 &
  46.8 &
  62.1 &
  \multicolumn{1}{c}{73.2} &
  58.6 &
  46.9 &
  66.8 &
  \multicolumn{1}{c}{79.5} &
  - &
  - &
  - &
  - \\
\multicolumn{1}{l}{BoxTE (2022)} &
  61.5 &
  53.2 &
  \bf 66.7 &
  \multicolumn{1}{c}{76.7} &
  66.7 &
  58.2&
  71.9 &
  \multicolumn{1}{c}{82.0} &
  - &
  - &
  - &
  - \\ 
\multicolumn{1}{l}{TeLM (2021)\footnotemark[2]} &
  61.41 &
  53.39 &
  66.0 &
  \multicolumn{1}{c}{76.12} &
  66.70 &
  58.99 &
  71.28 &
  \multicolumn{1}{c}{80.85} &
  - &
  - &
  - &
  - \\ 
  \midrule
\multicolumn{1}{l}{ChronoR (2021)\footnotemark[2]} &
  56.97 &
  46.50 &
  63.66 &
  \multicolumn{1}{c}{76.06} &
  60.64 &
  49.39 &
  67.97 &
  \multicolumn{1}{c}{81.13} &
  32.89 &
  25.88 &
  33.51 &
  50.71 \\
\multicolumn{1}{l}{ChronoR+Np/Lp (ours)} &
  57.45 &
  47.26 &
  64.00 &
  \multicolumn{1}{c}{76.06} &
  62.10 &
  51.62 &
  68.96 &
  \multicolumn{1}{c}{81.06} &
  33.89 &
  27.14 &
  34.33 &
  51.40 \\
\multicolumn{1}{l}{TNTComplEx (2020)\footnotemark[2]} &
  60.72 &
  51.91 &
  65.92 &
  \multicolumn{1}{c}{\bf 77.17} &
  66.64 &
  58.34 &
  71.82 &
  \multicolumn{1}{c}{81.67} &
  35.94 &
  28.49 &
  36.84 &
  53.75 \\
\multicolumn{1}{l}{TNTComplEx+Np/Lp (ours)} &
  \bf 61.80 &
  \bf 53.60 &
  66.55 &
  \multicolumn{1}{c}{76.97} &
  \bf 67.70 &
  \bf 59.90 &
  \bf 72.35 &
  \multicolumn{1}{c}{\bf82.30} &
  \textbf{37.05} &
  \textbf{29.00} &
  \textbf{39.62} &
  \textbf{54.02} \\
  \bottomrule
\end{tabular}
}
\caption[Evaluation on the YAGO15K, ICEWS14, and ICEWS05-15 datasets.]{Evaluation on the YAGO15K, ICEWS14, and ICEWS05-15 datasets. Results reported for previous related works are the best numbers reported in their respective paper.\footnotemark[2]
Results for our experiments on ChronoR and TNTComplEx (ours) are reported for the best configuration of hyperparameters based on the validation MRR.}
\label{tab:results}
\end{table*}

\begin{table*}[t]
\setlength{\tabcolsep}{4pt}
\resizebox{\textwidth}{!}{%
\begin{tabular}{lcccccccccccc}
\toprule
 &
  \multicolumn{4}{c}{{ICEWS14}} &
  \multicolumn{4}{c}{{ICEWS05-15}} &
  \multicolumn{4}{c}{{YAGO15K}} \\

\cmidrule(lr){2-5}
\cmidrule(lr){6-9}
\cmidrule(lr){10-13}
  
\multicolumn{1}{l}{Model} &
  MRR &
  Hit@1 &
  Hits@3 &
  \multicolumn{1}{c}{Hit@10} &
  MRR &
  Hit@1 &
  Hit@3 &
  \multicolumn{1}{c}{Hit@10} &
  MRR &
  Hit@1 &
  Hit@3 &
  Hit@10 \\
  \midrule
\multicolumn{1}{l}{ChronoR + N3 (2021)} &
  56.97 &
  46.50 &
  63.66 &
  \multicolumn{1}{c}{76.06} &
  60.64 &
  49.39 &
  67.97 &
  \multicolumn{1}{c}{81.13} &
  32.89 &
  25.88 &
  33.51 &
  50.71 \\
\multicolumn{1}{l}{ChronoR + L1} &
  56.91 & 
  46.45 &
  63.60 &
  \multicolumn{1}{c}{75.76} &
  53.89 &
  41.96 &
  60.65 &
  \multicolumn{1}{c}{76.82} &
  33.86 &
  27.04 &
  34.27 &
  51.33 \\
\multicolumn{1}{l}{ChronoR + L2} &
  57.42 &
  47.12 &
  63.97 &
  \multicolumn{1}{c}{76.35} &
  59.75 &
  48.99 &
  66.77 &
  \multicolumn{1}{c}{79.01} &
  \textbf{33.89} &
  \textbf{27.14} &
  34.33 &
  51.40 \\
\multicolumn{1}{l}{ChronoR + L3} &
  57.34 &
  46.94 &
  63.83 &
  \multicolumn{1}{c}{76.06} &
  \textbf{62.10} &
  \textbf{51.62} &
  \textbf{68.96} &
  \multicolumn{1}{c}{81.06} &
  32.98 &
  25.88 &
  33.75 &
  50.98 \\
\multicolumn{1}{l}{ChronoR + L4} &
  \textbf{57.45} &
  \textbf{47.26} &
  \textbf{64.00} &
  \multicolumn{1}{c}{76.06} &
  61.39 &
  50.53 &
  68.49 &
  \multicolumn{1}{c}{81.00} &
  33.31 &
  26.27 &
  33.97 &
  51.20 \\
\multicolumn{1}{l}{ChronoR + L5} &
  57.26 &
  46.96 &
  63.67 &
  \multicolumn{1}{c}{76.12} &
  62.02 &
  51.49 &
  68.87 &
  \multicolumn{1}{c}{81.07} &
  33.33 &
  26.27 &
  34.00 &
  51.34 \\
\multicolumn{1}{l}{ChronoR + N2} &
  56.89 &
  46.32 &
  63.71 &
  \multicolumn{1}{c}{76.14} &
  61.68 &
  50.87 &
  68.83 &
  \multicolumn{1}{c}{81.14} &
  33.08 &
  26.08 &
  33.57 &
  51.12 \\
\multicolumn{1}{l}{ChronoR + N4} &
  56.99 &
  46.39 &
  63.80 &
  \multicolumn{1}{c}{76.23} &
  60.95 &
  49.82 &
  68.26 &
  \multicolumn{1}{c}{\textbf{81.35}} &
  32.31 &
  24.25 &
  33.98 &
  51.62 \\
\multicolumn{1}{l}{ChronoR + N5} &
  57.01 &
  46.37 &
  63.82 &
  \multicolumn{1}{c}{\textbf{76.41}} &
  60.26 &
  49.00 &
  67.58 &
  \multicolumn{1}{c}{81.01} &
  33.03 &
  25.18 &
  \textbf{34.43} &
  \textbf{51.96} \\
  \midrule
\multicolumn{1}{l}{TNTComplEx + N3 (2020)} &
  60.72 &
  51.91 &
  65.92 &
  \multicolumn{1}{c}{\textbf{77.17}} &
  66.64 &
  58.34 &
  71.82 &
  \multicolumn{1}{c}{81.67} &
  35.94 &
  28.49 &
  36.84 &
  53.75 \\
\multicolumn{1}{l}{TNTComplEx + L1} &
  61.05 &
  52.92 &
  65.69 &
  \multicolumn{1}{c}{76.34} &
  65.38 &
  57.05 &
  70.58 &
  \multicolumn{1}{c}{80.45} &
  36.71 &
  28.75 &
  39.07 &
  53.64 \\
\multicolumn{1}{l}{TNTComplEx + L2} &
  61.40 &
  53.20 &
  66.27 &
  \multicolumn{1}{c}{76.65} &
  66.69 &
  58.63 &
  71.56 &
  \multicolumn{1}{c}{81.40} &
  36.88 &
  28.84 &
  39.42 &
  54.00 \\
\multicolumn{1}{l}{TNTComplEx + L3} &
  60.85 &
  52.24 &
  66.18 &
  \multicolumn{1}{c}{76.45} &
  65.58 &
  56.48 &
  71.27 &
  \multicolumn{1}{c}{\textbf{82.41}} &
  36.72 &
  28.71 &
  39.34 &
  53.45 \\
\multicolumn{1}{l}{TNTComplEx + L4} &
  61.34 &
  53.07 &
  66.16 &
  \multicolumn{1}{c}{76.51} &
  66.81 &
  58.49 &
  72.05 &
  \multicolumn{1}{c}{81.92} &
  36.87 &
  28.64 &
  \textbf{39.66} &
  54.25 \\
\multicolumn{1}{l}{TNTComplEx + L5} &
  61.36 &
  53.17 &
  66.11 &
  \multicolumn{1}{c}{76.56} &
  65.35 &
  56.42 &
  70.69 &
  \multicolumn{1}{c}{82.09} &
  36.84 &
  28.58 &
  39.46 &
  54.49 \\
\multicolumn{1}{l}{TNTComplEx + N2} &
  61.18 &
  52.87 &
  66.27 &
  \multicolumn{1}{c}{76.26} &
  66.96 &
  59.01 &
  71.79 &
  \multicolumn{1}{c}{81.56} &
  36.52 &
  29.00 &
  37.65 &
  54.39 \\
\multicolumn{1}{l}{TNTComplEx + N4} &
  \textbf{61.80} &
  \textbf{53.60} &
  \textbf{66.55} &
  \multicolumn{1}{c}{76.97} &
  67.59 &
  59.73 &
  72.26 &
  \multicolumn{1}{c}{82.20} &
  36.81 &
  29.24 &
  38.02 &
  \textbf{54.50} \\
\multicolumn{1}{l}{TNTComplEx + N5} &
  61.60 &
  53.41 &
  66.42 &
  \multicolumn{1}{c}{76.73} &
  \textbf{67.70} &
  \textbf{59.90} &
  \textbf{72.35} &
  \multicolumn{1}{c}{82.30} &
  \textbf{37.05} &
  29.00 &
  39.62 &
  54.02 \\
\multicolumn{1}{l}{TNTComplEx + RNN} &
  57.87 &
  47.95 &
  63.85 &
  \multicolumn{1}{c}{76.33} &
  54.35 &
  42.49 &
  61.38 &
  \multicolumn{1}{c}{76.94} &
  36.15 &
  \textbf{29.55} &
  36.42 &
  53.53 \\
\multicolumn{1}{l}{TNTComplEx + Linear3} &
  61.53 &
  53.31 &
  66.30 &
  \multicolumn{1}{c}{76.77} &
  67.54 &
  59.50 &
  72.53 &
  \multicolumn{1}{c}{82.19} &
  36.60 &
  28.38 &
  39.17 &
  54.14 \\
\midrule
\end{tabular}
}
\caption[Evaluation on the YAGO15K, ICEWS14, and ICEWS05-15 datasets.]{Evaluation on the YAGO15K, ICEWS14, and ICEWS05-15 datasets for TNTComplEx and ChronoR models using different temporal regularisers. We highlight the best results for both families of models.}
\label{tab:results-reg}
\end{table*}

\begin{figure}
    \centering
    \includegraphics[scale=0.3]{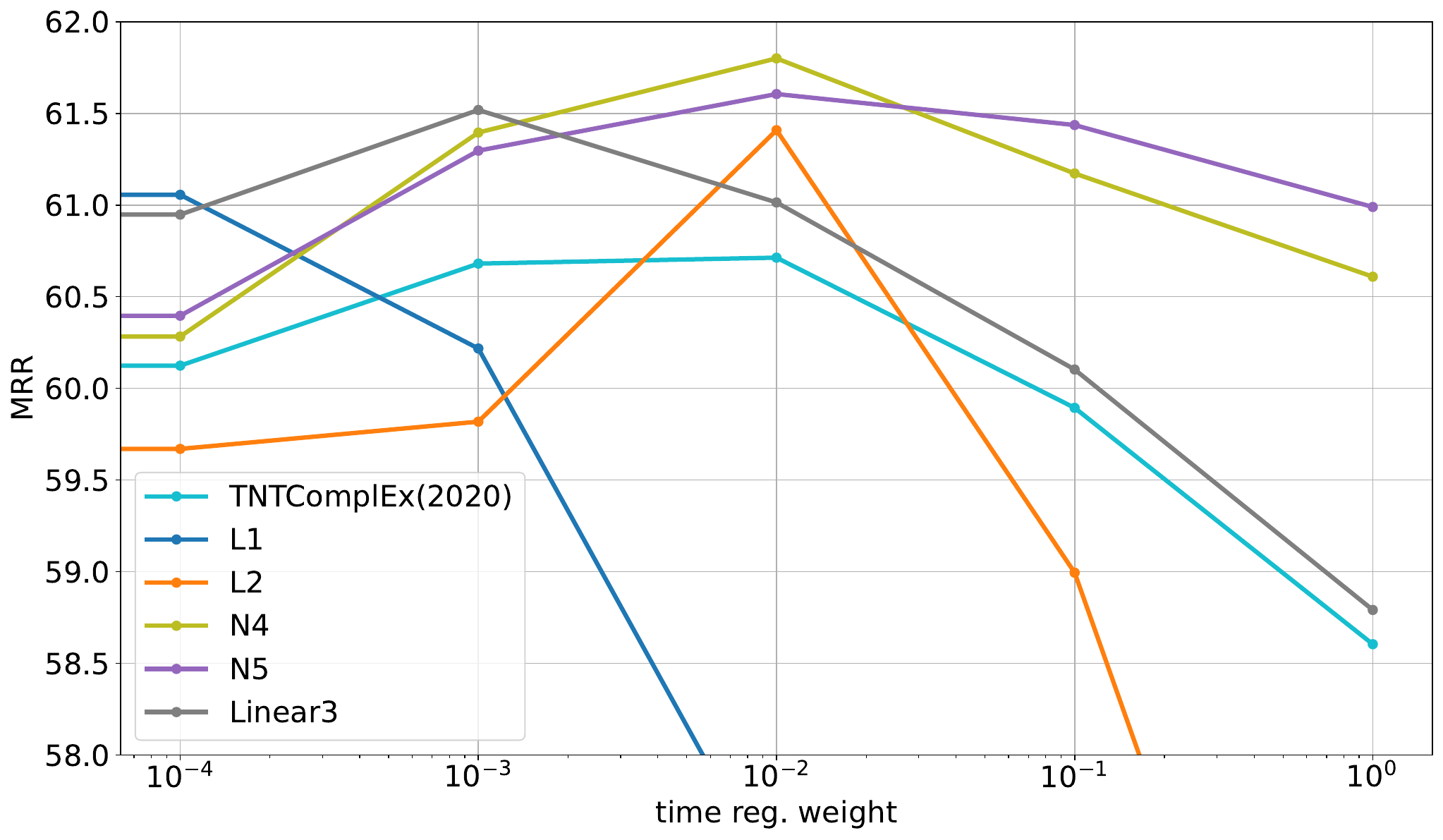}
    \caption{Comparison of various temporal regularisers with different regularisation weights on TNTComplEx trained on ICEWS14.}
    \label{fig:tntcomplex_icews14_timeregweight}
\end{figure}

We use baselines from both static and temporal KG embedding models.
We use TransE, DistMult, SimplE, and ComplEx from the static KG embedding models.
These models ignore the timing information. An important consideration when evaluating these models on temporal KGs in the filtered setting is that it is necessary for each test quadruple to filter out previously encountered entities based on the corresponding fact and its associated timestamp. This step ensures a fair comparison between different models.
To analyse the impact of temporal regularisers on tensor factorisation models, we compare our TNTComplEx and ChronoR models against their original counterparts. To the best of our knowledge, we also compare our results with all temporal KG embedding models from the literature that has been evaluated on these datasets, which we discussed the details of in \cref{sec:background}, except for TeMP \cite{DBLP:conf/emnlp/WuCCH20}, as it uses negative sampling for the evaluation, leading to an unfair comparison with our models.

\subsection{Results}
In this section, we analyse and perform a quantitative comparison of our models and previous state-of-the-art ones. We also experimentally verify the impact of several temporal regularisers on tensor factorisation models.

\cref{tab:results} demonstrates link prediction performance comparison on all datasets. Our TNTComplEx and ChronoR models achieve better performance than their original counterparts. Most importantly, TNTComplEx, with one of our proposed temporal regularisers, outperforms all the competitors in terms of link prediction MRR and Hits@1 metric on the three datasets.

\cref{tab:results-reg} shows link prediction performance comparison between all the considered temporal regularisers. The results show that the most accurate models on the three datasets are trained by setting the value of the hyperparameter $p$ to four or five, \ie the greatest choices considered. As described in \cref{sec:method}, the $p$ value determines the magnitude of the penalty and contributes to controlling the strength of the temporal smoothness. In L$_{p}$ norms, the penalty values decrease as $p$ increases, while in N$_{p}$ norms, larger values for $p$ widen the range of differences not to penalise. Hence, we achieve state-of-the-art performances on temporal knowledge graph completion on the three datasets by weakening the strength of temporal smoothness. It is important to recall that temporal smoothing improves the performance of tensor factorisation models \cite{DBLP:conf/iclr/LacroixOU20, DBLP:conf/aaai/SadeghianACW21, DBLP:conf/naacl/XuCNL21} so reduce its strength to zero (i.e. by removing its penalty term from the loss function) would not be beneficial.
\footnotetext[2]{The original results for TNTComplEx were reported on the validation set; we use the code and hyper-parameters from the official repository to re-run the model and report test set values.
The original implementation of ChronoR used by \citet{DBLP:conf/aaai/SadeghianACW21} is unavailable; we re-implement the solution in our framework, reporting the result according to the original configuration of hyperparameters.
TeLM uses approximately the same number of parameters as TNTComplEx and ChronoR; more information is available in the appendix.}
In ChronoR, the best performances are achieved using L$_{p}$ norms, while an opposite trend characterizes TNTComplEx. Therefore, it seems the scoring function adopted by the tensor factorisation models influences the choice of the temporal regulariser. In our intuition, the assumption made by ChronoR in Eq. \ref{eq:rotate}, which leads to its scoring function, imposes an additional constraint on the embedding representations, compared to TNTComplEx, that leads to benefit L$_{p}$ norms. %that leads timestamps representations to be closer in space. In this context, Lp norms, which always penalize no equal timestamp representations (see Figure \ref{fig:norms}), is a favourable choice.

Linear3 achieves the second-best results on all three datasets.
Indeed, the embedding representation $\bf{W_b}$ (see \cref{eq:linear3}) is adjusted through the training process so the models can learn how to weaken the temporal smoothing. On the contrary, using a recurrent architecture for temporal regularisation leads to the worst results. In our intuition, RNNs struggle to generate very long sequences of embeddings, as in our three datasets (see Timestamps in \cref{tab:stats}). Indeed, in our experiments, as the number of timestamps to generate increases, the output of the RNN at step $\tau$ converges to be equal to the output of step $\tau-1$.
In \cref{fig:tntcomplex_icews14_timeregweight}, we plot a detailed comparison of some of our proposed regularisers to the one proposed by \citet{DBLP:conf/iclr/LacroixOU20} for TNTComplEx, on ICEWS14 dataset. N$_{4}$ increases MRR by 1.08 points, and carefully selecting regularisation weight can increase MRR up to 3.2 points.

\section{Conclusion}
\label{sec:conclusion}
In this work, we analyse the impact of several temporal regularisers on tensor factorisation models for link prediction in temporal knowledge graphs. Specifically, we compare several choices of temporal smoothing regularisers using N$_{p}$ and L$_{p}$ norms, linear functions, and recurrent architectures.
Our experiments show that we can significantly improve the downstream link prediction accuracy in TNTComplEx and ChronoR by carefully selecting a temporal regulariser and corresponding weight hyperparameter.
By doing so, our version of TNTComplEx outperforms all baselines in terms of MRR on three benchmark datasets. Overall, temporal regularisers that weaken the temporal smoothing penalty on shorter time embedding differences, like N$_{4}$, N$_{5}$, and Linear3, produce the best results in terms of temporal link prediction accuracy across all considered datasets.

\paragraph{Future Works} We plan to extend our analysis to inductive tasks and show how temporal regularisers can be generalised to work with unseen timestamps, entities, and relation types -- for example, by leveraging recent work connecting factorisation-based models and GNNs~\citep{DBLP:conf/nips/ChenM0MS022}.
%

% For natbib users:
\bibliographystyle{unsrtnat}
\bibliography{reference}
% For bibLaTeX users:
% \printbibliography

\clearpage

\appendix

\section{Appendix}

\subsection{Description of Static Neural Link Predictors}
\paragraph{Scoring Functions}
Neural link prediction models can be characterised by their scoring function $\phi_{\theta}$.
For example, in TransE~\citep{DBLP:conf/nips/BordesUGWY13}, the score of a triple $\langle s, p, o \rangle$ is given by $\phi^{\text{TransE}}_{\theta}(s, p, o) = - \norm{\mathbf{s} + \mathbf{p} - \mathbf{o}}_{2}$, where $\mathbf{s}, \mathbf{p}, \mathbf{o} \in \mathbb{R}^{d}$ denote the embedding representations of $s$, $p$, and $o$, respectively.
In DistMult~\citep{DBLP:journals/corr/YangYHGD14a}, the scoring function is defined as $\phi^{\text{DistMult}}_{\theta}(s, p, o) = \langle \mathbf{s}, \mathbf{p}, \mathbf{o} \rangle = \sum_{z=1}^{d} \mathbf{s}_{z} \mathbf{p}_{z} \mathbf{o}_{z}$, where $\langle{}\cdot{}, {}\cdot{}, {}\cdot{} \rangle$ denotes the tri-linear dot product.
Canonical Tensor Decomposition~\citep[CP,][]{hitchcock-sum-1927} is similar to DistMult, with the difference that each entity $e \in \mathcal{E}$ has two representations, $\mathbf{e}_{\text{head}} \in \mathbb{R}^{d}$ and $\mathbf{e}_{\text{tail}} \in \mathbb{R}^{d}$, depending on whether it is being used as a head (subject) or tail (object): $\phi^{\text{CP}}_{\theta}(s, p, o) = \langle \mathbf{s}_{\text{head}}, \mathbf{p}, \mathbf{o}_{\text{tail}} \rangle$.
In RESCAL~\citep{DBLP:conf/icml/NickelTK11}, the scoring function is a bilinear model given by $\phi^{\text{RESCAL}}_{\theta}(s, p, o) = \mathbf{s}^{\top} \mathbf{P} \mathbf{o}$, where $\mathbf{s}, \mathbf{o} \in \mathbb{R}^{d}$ is the embedding representation of $s$ and $o$, and $\mathbf{P} \in \mathbb{R}^{d \times d}$ is the representation of $p$.
DistMult is equivalent to RESCAL if $\mathbf{P}$ is constrained to be diagonal.
Another variation of this model is ComplEx~\citep{DBLP:conf/icml/TrouillonWRGB16}, where the embedding representations of $s$, $p$, and $o$ are complex vectors -- \ie $\mathbf{s}, \mathbf{p}, \mathbf{o} \in \mathbb{C}^{d}$ -- and the scoring function is given by $\phi^{\text{ComplEx}}_{\theta}(s, p, o) = \Re(\langle \mathbf{s}, \mathbf{p}, \overline{\mathbf{o}} \rangle)$, where $\Re(\mathbf{x})$ represents the real part of $\mathbf{x}$, and $\overline{\mathbf{x}}$ denotes the complex conjugate of $\mathbf{x}$.
%
%In TuckER~\citep{Balazevic2019TuckERTF}, the scoring function is defined as follows:
%
%\begin{equation}
%$\phi^{\text{TuckER}}_{\theta}(s, p, o) = \mathbf{W} \times_{1} \mathbf{s} \times_{2} \mathbf{p} \times_{3} \mathbf{o}$,
%\end{equation}
%
%where $\mathbf{W} \in \mathbb{R}^{k_{s} \times k_{p} \times k_{o}}$ is a three-way tensor of parameters, and $\mathbf{s} \in \mathbb{R}^{k_{s}}$, $\mathbf{p} \in \mathbb{R}^{k_{p}}$, and $\mathbf{o} \in \mathbb{R}^{k_{o}}$ are the embedding representations of $s$, $p$, and $o$.
%

%
\paragraph{Training Objectives}
Another dimension for characterising neural link predictors is their training objective.
Early neural link prediction models such as RESCAL and CP were trained to minimise the reconstruction error of the whole adjacency tensor~\citep{DBLP:conf/icml/NickelTK11,DBLP:conf/acssc/VervlietDL16,DBLP:journals/jmlr/KossaifiPAP19}.
To scale to larger Knowledge Graphs, subsequent approaches such as \citet{DBLP:conf/nips/BordesUGWY13} and \citet{DBLP:journals/corr/YangYHGD14a} simplified the training objective by using \emph{negative sampling}: for each training triple, a corruption process generates a batch of negative examples by corrupting the subject and object of the triple, and the model is trained by increasing the score of the training triple while decreasing the score of its corruptions.
This approach was later extended by \citet{DBLP:conf/aaai/DettmersMS018} where, given a subject $s$ and a predicate $p$, the task of predicting the correct objects is cast as a $|\mathcal{E}|$-dimensional multi-label classification task, where each label corresponds to a distinct object and multiple labels can be assigned to the $(s, p)$ pair; this training objective is referred to as KvsAll by \citet{DBLP:conf/iclr/RuffinelliBG20}.
Another extension was proposed by \citet{DBLP:conf/icml/LacroixUO18} where, given a subject $s$ and a predicate $p$, the task of predicting the correct object $o$ in the training triple is cast as a $|\mathcal{E}|$-dimensional multi-class classification task, where each class corresponds to a distinct object and only one class can be assigned to the $(s, p)$ pair; this is referred to as 1vsAll by \citet{DBLP:conf/iclr/RuffinelliBG20}.
\paragraph{Regularisers}
As noted by \citet{DBLP:conf/nips/BordesUGWY13}, imposing regularisation terms on the learned entity and relation representations prevents the training process from trivially optimising the training objective by increasing the embedding norms.
Early works such as \citet{DBLP:conf/nips/BordesUGWY13,DBLP:conf/aaai/BordesWCB11,DBLP:journals/corr/abs-1301-3485,DBLP:conf/nips/JenattonRBO12} proposed constraining the embedding norms.
More recently, \citet{DBLP:journals/corr/YangYHGD14a,DBLP:conf/icml/TrouillonWRGB16} proposed adding a $L_{2}$ regularisation term on entity and relation representations to the training objective.
Lastly, \citet{DBLP:conf/icml/LacroixUO18} observed systematic improvements by replacing the $L_{2}$ norm with a nuclear tensor 3-norm. %, referred to as $N_{3}$.
%

%There has been a substantial amount of research in KG embedding in the non-temporal domain. One of the earliest models, TransE~\cite{bordes2013translating}, is a translational distance-based model, which embeds the entities \textbf{h} and \textbf{t}, along with relation \textbf{r}, and maps them through the function: \textbf{h} + \textbf{r} = \textbf{t}. 
%There have been several extensions to TransE, including TransH~\cite{wang2014knowledge} which models relations as hyperplanes; TransR~\cite{lin2015learning} which embed entities and relations in separate spaces and TransD~\cite{ji2015knowledge} in which two vectors represent each element in a triple in order to represent the elements and construct mapping matrices.
%Other works, such as DistMult~\cite{yang2014embedding}, represent relations as bilinear functions.  ComplEx~\cite{trouillon2016complex} extends DistMult to the complex space. RotatE~\cite{sun2019rotate} also embeds the entities in the complex space, and treats relations as planar rotations. QuatE~\cite{zhang2019quaternion} embeds each element using quaternions.

\subsection{Description of temporal knowledge graph completion models}

The subject of temporal link prediction has been studied in a wide range of approaches~\cite{DBLP:journals/jmlr/KazemiGJKSFP20}. One approach is to create a static KG by combining links from different time points without considering timestamps~\cite{DBLP:journals/jasis/Liben-NowellK07}. Afterwards, static embeddings are learned for individual entities. Some attempts have been made to enhance this method by assigning higher importance to more recent links~\cite{DBLP:conf/icdm/SharanN08,DBLP:journals/apin/IbrahimC15,DBLP:journals/isci/IbrahimC16,DBLP:journals/isci/IbrahimCWLLL16}. In contrast to these methods, \cite{DBLP:conf/ant/YaoWPY16} involves learning embeddings for each KG snapshot and then combining them using weighted averaging. Various techniques have been proposed for determining the weights of the embedding aggregation, including approaches based on ARIMA \cite{DBLP:journals/datamine/GunesOC16} or reinforcement learning \cite{MORADABADI2017422}.

Some additional studies employ sequence models for TKGs. \citet{DBLP:journals/jmlr/SarkarSG07} apply a Kalman filter to learn dynamic node embeddings. \citet{DBLP:conf/emnlp/Garcia-DuranDN18} extends DistMult and TransE to TKG using recurrent neural nets (RNN) for temporal data. For each relation, a temporal embedding has been learned by feeding a string version for timestamp (e.g. a date in US format) and the static relation embedding to an LSTM. This method learns dynamic embedding for relations but not for entities. Furthermore, \cite{han2020graph} employ a temporal point process parameterised by a deep neural architecture.

One of the earliest works that used representation learning techniques for reasoning over TKGs~\cite{sadeghian2016temporal} introduced an approach involving both embedding and rule mining methods for TKG reasoning. Another related model, t-TransE~\cite{DBLP:conf/emnlp/JiangLGSLCS16}, indirectly learns time-based embeddings by capturing the order of time-sensitive relations, such as \textit{wasBornIn} followed by \textit{diedIn}. In a similar vein, \citet{DBLP:conf/fusion/EstebanTYBK16} enforce temporal order constraints on their data by augmenting their quadruple $(s,p,o,\tau:\text{Bool})$, where $\text{Bool}$ indicates if the fact vanishes or continues after time $t$. However, the model's evaluation is limited to medical and sensory data.

Some approaches have attempted to apply diachronic word embeddings to the TKG problem, motivated by their success.~\cite{DBLP:conf/emnlp/Garcia-DuranDN18, DBLP:conf/emnlp/DasguptaRT18}.
Diachronic methods map every (node, timestamp) or (relation, timestamp) pair to a hidden representation. \citet{DBLP:conf/aaai/GoelKBP20} propose learning dynamic embeddings by masking a fraction of the embedding weights with an activation function of frequencies and \citet{DBLP:journals/corr/abs-1911-07893}~embed the vectors as a direct function of time. 

Other methods like \citet{DBLP:conf/esws/SchlichtkrullKB18, hanwww, DBLP:conf/emnlp/WuCCH20} leverage message passing graph neural networks (MPNNs) to learn structure-based entity representations. For instance, \cite{DBLP:conf/emnlp/WuCCH20} learns entity representation at every timestamp and then aggregates representations across all timestamps using a gated recurrent unit or self-attention encoder. Similarly to other models, the final entity representations can subsequently be used with a static KGE model such as ComplEx.

Finally, most of the proposed methods do not evolve the embedding of entities over time~\cite{DBLP:journals/ws/MaTD19, DBLP:conf/iclr/LacroixOU20, DBLP:conf/coling/XuNAYL20, DBLP:conf/aaai/SadeghianACW21, DBLP:conf/naacl/XuCNL21, DBLP:conf/aaai/MessnerAC22}. Instead, they learn the temporal behaviour by using a representation for time. For instance, \citet{DBLP:conf/iclr/LacroixOU20} perform tensor decomposition based on the time representation, while \citet{DBLP:conf/aaai/SadeghianACW21} learns a $d$-dimensional rotation transformation parametrized by relation and time, such that after each fact's head entity representation is transformed using the rotation, it falls near its corresponding tail entity. 

In this work, we focus on the latter methods as they achieve state-of-the-art results on most of the benchmark datasets~\cite{DBLP:conf/aaai/SadeghianACW21, DBLP:conf/naacl/XuCNL21}. Specifically, we systematically analyse a wide array of temporal regularisers to understand their impact on both performance and learning temporal behaviour.

\subsection{Details related to experimental evaluation and results}

\paragraph{Model parameter counts}
\cref{tab:model-params} reports the model parameter counts for TNTComplEx, ChronoR and other competing models. For all the models, $d$ is the embedding size, $|E|$, $|T|$, and $|R|$ are the number of entities, timestamps and relations in the dataset, respectively. Note that ChronoR and TNTComplEx have the same number of parameters, while TeLM, which uses multi-vector embeddings, has exactly double.
 \begin{table}[t!]
    \centering
    \begin{tabular}{lc}
        \toprule
        Model & Number of Parameters\\
        \midrule
        DE-SimplE       & $2d((3\gamma + (1 - \gamma)) \left| \mathbf{E} \right| + \left| \mathbf{R} \right|)$ \\
        TComplEx        & $2d(\left| \mathbf{E} \right| + \left| \mathbf{T} \right| + 2 \left| \mathbf{R} \right|)$ \\
        BoxTE  & $d(2\left| \mathbf{E} \right| + k \left| \mathbf{T} \right| + 2 \left| \mathbf{R} \right|) + k \left| \mathbf{R} \right|$ \\
        TeLM      & $4d(\left| \mathbf{E} \right| + \left| \mathbf{T} \right| + 4 \left| \mathbf{R} \right|)$ \\ 
        ChronoR      & $2d(\left| \mathbf{E} \right| + \left| \mathbf{T} \right| + 4 \left| \mathbf{R} \right|)$ \\ 
        TNTComplEx      & $2d(\left| \mathbf{E} \right| + \left| \mathbf{T} \right| + 4 \left| \mathbf{R} \right|)$ \\
        \bottomrule
    \end{tabular}
    \caption{Model parameter count for TNTComplEx and other competing models. For DE-SimplE, $\gamma$ denotes the share of temporal embedding features. For BoxTE, $k$ denotes the dimension for the scalar vector.}
    \label{tab:model-params}
\end{table}

\paragraph{Results of previous related works.}

The results of previous related works are the best numbers reported in their respective paper excpet for TNTComplEx, ChronoR and TeLM. 
The original results for TNTComplEx were reported on the validation set, we use the code and hyper-parameters from the official repository to re-run the model and report test set values. The same approach to report TNTComplEx results was adopted by \cite{DBLP:conf/aaai/SadeghianACW21}.
The original implementation of ChronoR (2021) is not available, we re-implement the solution in our framework, reporting the result according to the original configuration of hyperparameters stated in their paper.
TeLM is trained using almost the same number of parameters as TNTComplEx and ChronoR. As shown in \cref{tab:model-params}, fixed an embedding dimension $d$, TeLM has double the number of parameters of TNTComplEx, hence we report the results of TeLM setting $d=1000$. In its original paper \cite{DBLP:conf/naacl/XuCNL21}, the results refer to $d=2000$.

\paragraph{Best configuration of hyperparameters} 

We tune all the hyper-parameters using a grid search using the validation set provided with each of the datasets.
We tune $\lambda_1$ and $\lambda_2$ from $\{10^i| -4 \leq i \leq 1\}$, the $p$ value for Np and Lp norms from $\{i | 1 \leq i \leq 5\}$, the embedding dimensions for tensors and recurrent architecture from $\{5, 25, 50, 100, 500, 2000\}$. 

The training was done using mini-batch stochastic gradient descent with Adam and a learning rate of 0.1 with a batch size of 1000 quadruples.

We report the best configuration of hyperparameters for TNTComplEx in Table \ref{tab:hyperparams}.
\begin{table*}
    \centering
    \begin{tabular}{lcccc}
        \toprule
        {\bf Dataset} & {\bf Embedding Dimension} & $\lambda_1$ & $\lambda_2$ & {\bf Temporal Regulariser} \\
        \midrule
        ICEWS14 & 2000 & 0.001 & 0.01 & N4\\
        ICEWS05-15 & 2000 & 0.001 & 1 & N5\\
        YAGO15K & 2000 & 0.0001 & 0.0001 & N5\\
        \bottomrule
    \end{tabular}
    \caption{Best hyper-parameter configuration of TNTComplEx on ICEWS14, ICEWS5-15, and YAGO15K -- $\lambda_{1}$ and $\lambda_{2}$ denote the N3 and the temporal regularisation weights, respectively.}
    \label{tab:hyperparams}
\end{table*}

\end{document}